# Beyond XAI:Obstacles Towards Responsible AI

**Yulu Pi**[*]
Centre for Interdisciplinary Methodologies
University of Warwick
Coventry, UK
yulu.pi@warwick.ac.uk

## Abstract

The rapidly advancing domain of Explainable Artificial Intelligence (XAI) has sparked significant interests in developing techniques to make AI systems more transparent and understandable. Nevertheless, in real-world contexts, the methods of explainability and their evaluation strategies present numerous limitations. Moreover, the scope of responsible AI extends beyond just explainability. In this paper, we explore these limitations and discuss their implications in a boarder context of responsible AI when considering other important aspects, including privacy, fairness and contestability.

## 1 Introduction

The barrier of explainability has spurred significant concerns as Artificial Intelligence (AI) lies at the core of many sectors. Explainable AI (XAI) emerged as a response to this, striving to make AI "behavior more intelligible to humans by providing explanations[20]." However, as we delve deeper into the application of XAI in real-world settings, it becomes evident XAI does not suffice on its own[9]. Explainability, while crucial, is merely one facet of the broader challenge. The quest for Responsible AI demands a more holistic perspective that surpasses the bounds of XAI.

Moving our vision beyond the XAI realm toward Responsible AI, we recognize the pressing need to address broader dimensions including privacy, fairness, accountability, and contestability. It is becoming increasingly clear that relying solely on XAI, without addressing these interwoven complexities, falls short of achieving Responsible AI. In this paper, we discuss the limitation of XAI from machine learning and human-computer interaction persepctives. We also explore the implications of adopting XAI techniques in the context of considering other responsible AI. Given our space constraints, we specifically hone in on fairness, privacy, and contestability.

## 2 Technical challenge of AI Explainability

ML/AI researchers mainly focus on developing methods that make the decision process of a model, less of a black-box. XAI methods can be divided into interpretable models designed from the start and post-hoc explanations derived from black-box models. Advocates of interpretable models argue that for high-stake decisions, prioritizing inherently transparent models is crucial[37]. Nevertheless, the application of black-box models is still the dominant practice, and there are still significant computational and technical hurdles in designing interpretable models, such as distilling a complex data space into an optimal set of discretized and meaningful features[7]. A common belief is that the black-box methods generally obtain higher accuracy than the interpretable ones[9]. As post-hoc methods are applied after the model's training process, they can be used to provide explainability in complex ML models without loss of performance[32].

---

[*]nikipi.github.io

Post-hoc methods can be categorized based on several dimensions[40], including the type of data they analyze[2] (e.g., tabular, text, image), the algorithm used[11] (e.g., differentiable or non-differentiable algorithms), the scope of explanations[19] (e.g., local explanations focusing on specific input-output relationships or global explanations addressing overall model behavior), the format of the explanation[31] (e.g., verbal, visual and interative interface), and the approach to explaining[42] (e.g., feature-based explanations for important factors in the decision-making process or contrastive reasoning involving similar and counterfactual example)

Despite the proliferation of XAI techniques, developing faithful and reliable explanations for various machine learning models remains one of the unsolved challenges. Much of existing work in XAI produces simplified approximations of complex original models, which result in different level of faithfulness, also refer as fidelity. Through experiments, including model parameter randomization and data randomization tests, [4] discovered that numerous XAI methods are incapable to generate explanations that truly reflect the model' logic or the data generating process. Furthermore, recent research discovered that explanations can be susceptible to manipulation. By Applying visually imperceptible pertubations to the input image that keep the network's output approximately constant,one can manipulate generated explanations arbitrarily[14]. Similarly, [39] demonstrate the possibility of modifying biased classifiers can easily fool popular explanation techniques such as LIME and SHAP into generating innocuous explanations which do not reflect the underlying biases. It raises concerns about the reliability of explanations and the possibility of malicious actors misleading users about how a model operates. They recommend using adversarial manipulation to evaluate the robustness and reliability of explanations[21]. The imperfection and vulnerabilities of technical approach of explainability complicates its evaluation and application in real-work context.

Moving towards responsible AI, the focus broadens beyond mere explainability. The term "responsible AI" is intended to encapsulate a broad set of technical and socio-technical attributes of AI systems such as safety, efficacy, fairness, privacy, transparency and explainability[41, 9]. There is growing awareness that expliability is intricately connected with other facets of Responsible AI. Yet, limited research studies explainability from a lens of its integration, support, and potential clashes with other essential facets of responsible AI. In the following paragraphs, we only explore how current research addressed explainability alongside fairness and privacy given the paper's length constraints. We advocate for more comprehensive and in-depth research to understand the intricacy of explainability for a boarder objective of responsible AI.

## 2.1 Fairness

Many research and AI ethical guideline emphasized the important relationship between explainability and fairness from a theoretical point of view, highlighting the instrumental role of explanations plays in AI fairness[46]. For instance, 26 out of the 28 AI principles surveyed that address XAI, also talk about fairness explicitly emphasizing both aspects together when implementing Responsible AI[18]. From a societal viewpoint, explainability is seen as a mechanism to ensure and champion fairness in AI[9]. In their seminal study, researchers underscore that explanations should serve as tools for humans to discern if AI decisions harbor biases against protected groups[15].

Feature-based XAI techniques such as SHAP, decompose the model output into feature attributions. This breakdown can be used to compute the quantitative fairness metric such as demographic parity difference for each input feature using the SHAP value. Such an examination via XAI can help detect implicit connections between protected and unprotected features[28]. Conversely, [24] suggests that explanations methods that generate counterfactual explanations of positive and negative evidence of fairness offer tangible value to those at the receiving end of an AI model's decisions. Such explanations rather than feature importance and actionable recourse, present evidence potentially pointing to historical instances of unfairness.

While there is a consensus on the crucial role of explainability in AI fairness, numerous critics argue that many XAI techniques fall short in providing essential functionalities for bias detection. In a survey examining the use of explainability methods to uncover or investigate biases in NLP, the majority of the works they identified utilized feature attribution methods[10].However, [39] proved that feature attribution techniques such as LIME and SHAP can be manipulated to hide the underlying biases of a biased model, leading to false belief of fairness. Additionally, there are inherent conceptual issues with such approaches. XAI techniques often check whether models recognize features tied to protected attributes instead of ensuring that the input data is devoid of biases concerning those attributes. This mirrors the "fairness through unawareness" strategy[10]. Yet, the role of XAI in working with other bias mitigation strategies, such as pre-processing, in-processing, and post-processing, has yet to be clarified.

[5] underscore six important functions, yet to be fully realized, of XAI in solving biased data and issues involved in the selection and formulation of ML models:

- The XAI tools could identify imbalances within the data as it relates to over/under-sampling;
- The XAI tools could identify attributes most influential in both local and global decisions;
- The XAI tools can identify processing issues that had a distinct impact on the final model;
- The XAI tools can consider the impact of user-labeled sensitive attributes on the model performance.
- The XAI tools can highlight influences from model selection and optimization that impacted the final algorithm and it's performance and
- The XAI tools consider some metric of fairness in evaluating the global performance of the resulting algorithm.

### 2.2 Privacy

Although explainability and privacy have been widely studied as two separate fields in previous research[27], tension and complexity between explainability and privacy have received growing concerns. Explainability strives to disclose more details about specific decisions or the overarching decision-making process, while privacy-centric techniques deliberately avoid in-depth revelations about individual decision path, concentrating on dataset-wide statistics. Providing additional details about the model for enhanced explainability may compromise privacy or vice versa. For instance, if images are obfuscated for privacy reasons, providing explanations for the classification may unintentionally reveal the identities of people in the images[45]. Furthermore, different explainability methods also have various level of privacy issues. Previous research showed that the success rate of attacks exploiting model explanations, especially backpropagation-based methods rather than perturbation-based methods surpassed those that only have access to model predictions[16].[38] conducted an extensive experimental analysis to understand the impact of private learning techniques on generated model explanations. Their research indicates complex interactions between privacy-perserving and explainability techniques. For instance, differential Privacy methods hampers the interpretability of explanations, while federated Learning often improve the understanding of generated explanations.

A deeper understanding for the intricate relationship between privacy and explainability especially critical for sensitive domains like medicine which pose growing demands on AI systems that balance data privacy with appropriate explainability. [27] investigated the privacy risk of explaination in the context of using AI for biomedical image analysis and found that differential privacy negatively influence the computation of concept-based explanations. Their research identified a need for an extra training procedure for differentially privacy for concept-based explanations. In addition to technical advancements, how to strike balance between explainability and privacy in different context requires a deeper analysis of the effect of privacy on the explanation by human users through application-grounded and human-grounded evaluation methods.

## 3 Ambiguity of Explainability's Role in Human-AI relationships

Technical approaches to XAI have been criticized for their algorithm-centric focus, solely emphasizing the development of technical and mathematical methods to explain the behavior of underlying ML models[30, 3]. Functionally-grounded evaluations[15] of XAI methods rely on formal definitions of explainability such as fidelity and sensitivity[44] as a proxy to evaluate the quality of generated explanations without involving human experimentation. However, this evaluation method fails to consider essential human factors, such as the knowledge, needs, and expectations of real users[23, 25]. As a result, it does not provide measurements of explanation quality in the areas of improving trust, enhancing understanding, and facilitating performance of human and AI teams[30, 3].

Aware of this shortfall, many HCI researchers applied human-grounded evaluation, where XAI methods are evaluated with user studies but with simplified tasks[13, 43]. By conducting user studies, HCI aim to understand 'what makes a good explanation' by asking for whom the explanation is for(who), which is the goal of the AI explanation(why), and what the types of explanations are(how)[17]. While there are no definitive answers so far, HCI research reveals substantial variation in users' perceptions and attitudes towards explainability. Its role in enhancing understanding, fostering trust, and facilitating human-AI collaboration appears to shift significantly depending on the task given and application context[43, 26].

HCI research further underscores that different stakeholders require various types of explanations based on their specific purposes[23], the domain of application, and a plethora of human factors, such as AI literacy and cultural background[33]. In varying contexts and for distinct users, the desirable qualities of explainability satisfy different, sometimes conflicting. Take for instance the "faithfulness" feature, which assesses if the explanation accurately reflects the inner workings of the complex model, versus the "compactness" feature, ensuring the explanation remains concise and not overwhelming[25]. An ML engineer may prioritize faithful explanations for model debugging and development, while an end user might prefer concise explanations, even if they aren't as detailed[34].

However, these human and contextual factors have not been fully understood yet, with different studies presenting varied findings. For example, Cheng et al's work provided concerted empirical evidence that interactive visual explanations are effective at improving non-expert users' comprehension of algorithmic decisions[13]. Conversely, Han Liu et al observed mixed results for interactive explanations: while these explanations improve human perception of AI assistance's usefulness, they may reinforce human biases and lead to marginal performance improvement[26]. This variation in findings points to a significant gap in HCI research: simple experiments can not to fully capture the complexities of real-world situations, failing to provide generalized guidance on XAI applications. Human-grounded evaluations frequently rely on proxy tasks with simplified experimental settings, making it hard to gauge how these explanations would help users in real-work tasks[8]. Given the evolving emphasis of explainability in AI regulations, there's an urgent need for more application-grounded evaluations where XAI methods are assessed in real-world scenarios with actual users. In real work settings, AI operates in a boarder context where procedural information matters when people try to understand AI.

Without knowing how different types of explanations change what users know, thereby enabling them to act in response to AI, any enforcement on providing explanations is unlikely to meet its goal. The lack of universally applicable or context-specific research findings introduces a significant complexity into the process of creating well-rounded policy recommendations or regulatory decisions. In other words, without research outcomes that can be applied broadly across various contexts or those tailored to specific situations, it becomes exceedingly challenging to design policies or regulations that effectively govern the use of AI. We have recognized "contestability" as a crucial aspect that explanations can enable, yet HCI does not adequately study due to the lack of real world evaluation. A detailed discussion on this matter follows below.

### 3.1 Contestability

In the context of automated decision-making, particularly customer-facing applications, there is a power imbalance where decision makers are in a position of power over decision subjects. The opacity of AI decision-making systems, either due to the mere complexity or proprietary claims, tend to exacerbate the existing power gap —decision makers have access to more information about the AI system that is not available to decision subjects[29]. Recently, scholars have begun to explore the right to contestability, the ability to contest algorithmic decisions. Many argue that contestability offers decision subjects some protection, permitting them to reclaim some control and hold decision-makers accountable. This often involves requesting a review, many times involving human intervention and scrutiny[41]. Contestability enables interactions among decision-makers and algorithms, decision-maker and system designers, and ideally between decision-makers and individuals impacted by the decisions, as well as the general public[22]. There is existing legislation that offers affected individuals the possibility to seek such review. For example, under the Equal Credit Opportunity Act, if a candidate's credit application is rejected, the credit bureau is obligated to share the main reasons for the rejection, thereby enabling the possibility of a contestation. While contestability empowers individuals to exercise their autonomy for their own advantage, some critics argue that it imposes an undue burden on those affected by the decision: *the onus is on the individual to pursue an appeal*[29].

However, the applicability of this regulatory approach to automated decision-making is not entirely clear. The challenges don't only lie in the technical difficulties of approximating the decision rules of these "black box" system. Automated decision-making also opens up new questions about defining what can be contested, who can contest, who is accountable and how the review process should be conducted[29]. Many have discussed the relationship between explainability and contestability. Providing an explanation is not only seen as complementary to contestability, but as an essential prerequisite that enables a person to contest a decision[6, 29, 35]. The importance of an explanation in determining whether the algorithmic decision was justified and to provide grounds for review have been underscored. While there is an agreement that explanations should contain the information necessary for a decision subject to exercise their rights contestation[12, 36], defining what constitutes adequate and relevant information remains a complex issue. UK government calls for more evidence on interactions with requirements of appropriate explainability, acting as pre-conditions of effective redress and contestability in its recent white paper on AI regulation[1].

Research in XAI has increasingly shifted towards more human-centred approaches in designing and evaluating explanations. The provision of explanations specifically for contestation necessitates a well-defined decision-making context, which current proxy tasks often fail to deliver. The question of what explanations and interactions will prompt appropriate engagement and contestation by impacted individuals is domain- and context-specific[35]. In the rising landscape of AI governance, there's a growing necessity for XAI contributions that consider contestation as a vital factor to inform policy decisions.

## 4 Closing Remarks

In this paper, we explore the limitations of current explainability methods and evaluation practices, as well as the implications when considering other important factors for responsible AI, including privacy, fairness, and contestability. Our exploration has shown that while explainability is undeniably pivotal, the pursuit of responsible AI practices is hindered without a well-established mechanism linking explainability to other critical aspects.

In sum, while explainability is a commendable stride forward, the ultimate destination is a comprehensive and holistic approach to responsible AI—one that embraces and integrates all its multifaceted dimensions.